\documentclass[a4paper,twoside]{article}

\usepackage{epsfig}
\usepackage{subcaption}
\usepackage{calc}
\usepackage{amssymb}
\usepackage{amstext}
\usepackage{amsmath}
\usepackage{amsthm}
\usepackage{multicol}
\usepackage{pslatex}
\usepackage{apalike}
\usepackage{algorithm2e}
\usepackage[bottom]{footmisc}
\usepackage{SCITEPRESS}     

\usepackage{amsmath,amssymb,amsfonts}
\usepackage{algorithmic}
\usepackage{graphicx}
\usepackage{textcomp}
\usepackage{xcolor}

\usepackage[utf8]{inputenc} 
\usepackage[T1]{fontenc}    
\usepackage[hidelinks]{hyperref}       
\usepackage{url}            
\usepackage{booktabs}       
\usepackage{amsfonts}       
\usepackage{nicefrac}       
\usepackage{microtype}      
\usepackage{cite}

\begin{document}

\title{Evaluating Data Augmentation Techniques for Coffee Leaf Disease Classification}

\author{\authorname{Adrian Gheorghiu\sup{1}\sup{*}, Iulian-Marius Tăiatu\sup{1}\sup{*}, Dumitru-Clementin Cercel\sup{1}\thanks{Corresponding author.}, 
Iuliana Marin\sup{2},  \newline and Florin Pop\sup{1,3,4}}
\affiliation{\sup{1} Computer Science and Engineering Department, National University of Science and Technology POLITEHNICA Bucharest}
\affiliation{\sup{2}  Faculty of Engineering in Foreign Languages, National University of Science and Technology POLITEHNICA Bucharest}
\affiliation{\sup{3} National Institute for Research and Development in Informatics - ICI Bucharest, Romania}
\affiliation{\sup{4} Academy of Romanian Scientists, Bucharest, Romania}
\email{\{adrian.gheorghiu00, iulian.taiatu\}@stud.acs.upb.ro, \{dumitru.cercel, iuliana.marin, florin.pop\}@upb.ro}
}


\abstract{The detection and classification of diseases in Robusta coffee leaves are essential to ensure that plants are healthy and the crop yield is kept high. However, this job requires extensive botanical knowledge and much wasted time. Therefore, this task and others similar to it have been extensively researched subjects in image classification. Regarding leaf disease classification, most approaches have used the more popular PlantVillage dataset while completely disregarding other datasets, like the Robusta Coffee Leaf (RoCoLe) dataset. As the RoCoLe dataset is imbalanced and does not have many samples, fine-tuning of pre-trained models and multiple augmentation techniques need to be used. The current paper uses the RoCoLe dataset and approaches based on deep learning for classifying coffee leaf diseases from images, incorporating the pix2pix model for segmentation and cycle-generative adversarial network (CycleGAN) for augmentation. Our study demonstrates the effectiveness of Transformer-based models, online augmentations, and CycleGAN augmentation in improving leaf disease classification. While synthetic data has limitations, it complements real data, enhancing model performance. These findings contribute to developing robust techniques for plant disease detection and classification.}

\onecolumn \maketitle \normalsize \setcounter{footnote}{0} \vfill

\def\thefootnote{*}\footnotetext{Equal contributions.}\def\thefootnote{\arabic{footnote}}


\section{Introduction}
\label{sec:intro}

The Robusta coffee plant (also called Coffea Canephora) is a species of coffee that is susceptible to many diseases. Whether those diseases are caused by insects or fungi, they can have a major impact on crop yields and even cause complete crop destruction if left untreated. The detection of diseases from images has been a significant focus in the research of classification tasks for many years, starting with diseases in humans  \cite{ding2023multi, sun2023instance, lungu2023skindistilvit} and then moving to animals \cite{stauber2008maldi, naas2020infrared, nam2023classification} and plants \cite{dawod2022automatic, dawod2022resnet, dawod2022upper, echim2023explainabilitydriven}.

Generally, detecting diseases requires expert knowledge and a lot of time spent analyzing images to determine the severity of the disease. For this reason, there have been many research interests in developing machine learning tools that anyone can use to detect and classify diseases \cite{kamal2019depthwise, dawod2021classification}. These tools must be easy to use, accurate, and not overly confident when making wrong predictions.

In this paper, we use the Robusta Coffee Leaf (RoCoLe) dataset \cite{rocole} for which the main issues are the small number of available images and the class imbalance. These are both widespread problems that arise in the field of machine learning and for which multiple approaches exist. Therefore, several methods were tested and compared to determine which solution fits the RoCoLe dataset the best, with the main contributions of our work being as follows:

\begin{itemize}
    \item We test offline and online augmentations of the dataset conjointly with different combinations of models and hyperparameters. The main focuses of these comparisons are the performance evaluations of the augmentations, followed by the assessment of Transformer \cite{vaswani2017attention}-based models compared to a state-of-the-art convolutional model. 
   \item We employ different visualization and explainability techniques \cite{cam, hinton2002tsne} to better understand why the models perform in certain ways.
    \item To the best of our knowledge, we are the first to augment the RoCoLe dataset and use it for training and testing Transformer-based models.
\end{itemize}

\section{Related Work}
\label{sec:related}

Most leaf disease classification approaches used large datasets comprising tens of thousands of images \cite{plantvillage, thakur2023vgg}. Only some works used the RoCoLe dataset; if they do, it is not the primary training dataset and is mainly used for evaluation \cite{semantic_segmentation, rodriguez2023robust, faisal2023model}. The approaches used vary, with both deep learning and classical machine learning models being employed \cite{tomato_diseases, semantic_segmentation}.

Therefore, Brahimi et al. \cite{tomato_diseases} used a deep learning approach by testing two traditional convolutional neural network (CNN) \cite{kim2014convolutional} architectures: AlexNet \cite{alexnet} and GoogLeNet \cite{szegedy2015going}. Traditional machine learning models, specifically support vector machines (SVMs) and Random Forest, are also tested. The work compares pre-trained models to models without pre-training, as well as deep models trained on raw data to shallow models trained on manually extracted features. Feature activation visualization is also used as a more rudimentary technique. 
For this, regions of the image are sequentially occluded, after which the classification model is used to get the prediction. The negative log-likelihood is then used to estimate the importance of the occluded region. This method has the disadvantage of inefficiency and is only feasible when generating extremely low dimensional activation maps comprised of only a few pixels. For the SVM and Random Forest shallow models, the images are first transformed from the RGB color space to another color space, after which manual features such as the color moment, wavelet transform, and gray level co-occurrence matrix are extracted and used in training the classifiers.

Tassis et al.\cite{semantic_segmentation} used a multi-stage pipeline of three different models to segment and classify the leaf diseases. Thus, a Mask R-CNN model \cite{he2017mask} is first used for instance segmentation. This model is trained to mask the background and only highlight the leaves in images. After that, either a U-Net \cite{unet} or a PSPNet \cite{pspnet} is used for semantic segmentation. This model highlights the relevant areas of the studied image, specifically diseased regions. Those regions are then cropped and fed into a ResNet classifier \cite{resnet}, and the predictions are used to estimate the severity of the disease. The models were trained using random rotations and color variations as augmentations. As a training dataset, the authors used images scraped from the web. The RoCoLe dataset was also used to evaluate the instance segmentation model.

Mohameth et al. \cite{plantvillage} employed the popular PlantVillage dataset \cite{datplantvillage} to train their proposed models. For training, both transfer learning and deep feature extraction methods are used. For transfer learning, only the classification head of a pre-trained model is trained from scratch, with the rest of the layers having their weights frozen. Using this method, the authors test multiple CNN-based models: VGG16 \cite{simonyan2014very}, GoogLeNet, and ResNet. When deep feature extraction is utilized, these models only extract the features before the classification head. Those features are then used to train an SVM or a k-nearest neighbor classifier.



\section{Dataset}

\subsection{Relabeling}

By performing exploratory data analysis on the RoCoLe dataset, it becomes evident that the classes are imbalanced. This is a significant problem for any classifier architecture, as the model will overfit the most frequent class and rarely predict the less frequent classes.

It is easy to notice that the labels corresponding to the two most severe cases of rust are also the least frequent. The rust\_level\_3 and rust\_level\_4 labels both represent high levels of rust, and even by combining them, they still make up fewer samples than any other label. Therefore, the first step towards alleviating class imbalance is to relabel the samples as follows: healthy and red\_spider\_mite stay the same, rust\_level\_1 becomes rust\_level\_low, rust\_level\_2 becomes rust\_level\_medium, and both rust\_level\_3 and rust\_level\_4 become rust\_level\_high.

\subsection{Preprocessing}

Before further addressing the class imbalance, the next step was to analyze the images from the RoCoLe dataset. We notice that the photos were taken with a smartphone camera and, therefore, have a high resolution, 1152x2048, to be more precise. Using images at this resolution for deep learning models would consume many computational resources without the models benefiting much from the increased resolution. Therefore, the images were rescaled to the less demanding and more common 256x256 resolution. The images in the dataset also come with a mask, represented by the points that comprise the mask polygon. This mask was plotted, rescaled, and saved with its related image.

\subsection{Split}

\begin{figure}[!ht]
\centering
\includegraphics[width=1.0\columnwidth]{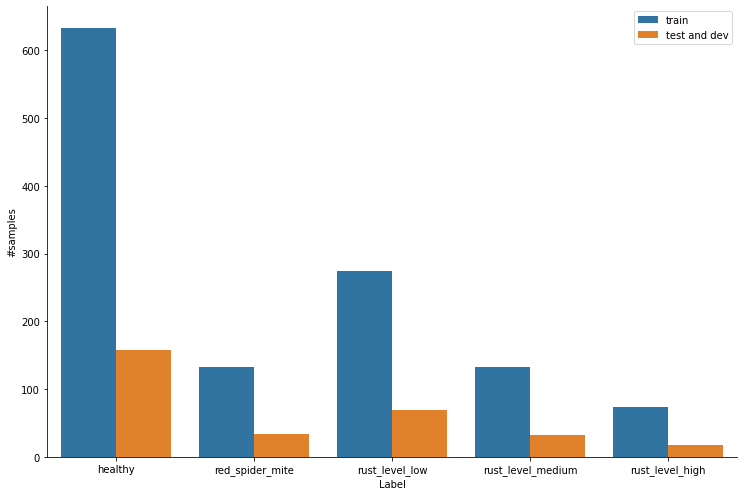}
\caption{Statistics of the number of samples in the relabeled dataset.}
\label{fig:statistics}
\end{figure}

The initial dataset is randomly split into train, development (dev), and test sets as shown in Figure \ref{fig:statistics}. The split is done as follows: 80\% goes to the train set, with the rest of 20\% being split equally between the dev and test sets. After the split, the train and dev sets are augmented together and then split back with an 80\%-10\% ratio. We do not augment the test set since the model metrics computed on the test set must only reflect the performance on data from the real input distribution.

\section{Method}
\label{sec:method}

\subsection{Segmentation}

\begin{figure}[!ht]
\centering
\includegraphics[width = 210pt, height = 130pt]{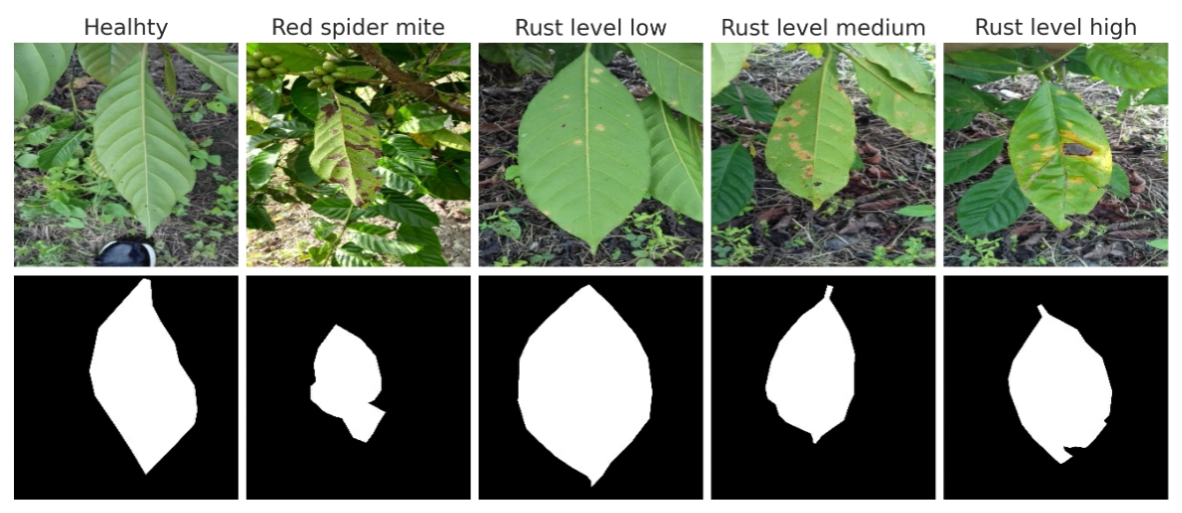}
\caption{Examples of rescaled images from each class and the associated masks.}
\label{fig:examples}
\end{figure}

As can be seen from the examples in Figure \ref{fig:examples}, the backgrounds of the images are very random and do not provide any additional information. Therefore, it only makes sense to use masked images to train classifiers. For training a classifier, it is trivial to apply the mask that comes with the image. Unfortunately, the issue is for the examples that do not already come segmented. The segmentation problem involves learning a mapping from an input distribution to the distribution that represents the mask of the inputs. Because of the pairwise nature of the segmentation problem, the pix2pix model \cite{pix2pix} is the perfect fit for it. Furthermore, pix2pix is based on the U-Net architecture, which is state of the art in image segmentation \cite{siddique2021u}. The pix2pix model was thus trained on image-mask pairs from the RoCoLe dataset. 

To improve the quality of the predicted segmentation masks, online augmentation was also used to crop and flip the image randomly. After training, the mask was inferred for each image in the dataset and then applied to the image to only contain the area of interest, namely the leaf in the foreground. The reason segmentation of the dataset is done using the trained pix2pix model and not using the already provided masks is because the inferred masks are slightly different from the provided ones. Training a classifier on images segmented with the provided masks but segmenting the test images with pix2pix will affect the quality of the predictions as the input distribution for the classifier will be slightly different.

\subsection{Offline Augmentations}

In order to alleviate the class imbalance, image generation was used to supplement the less frequent classes, thus making the dataset perfectly balanced. The class frequencies show that the healthy class is the most frequent. This means that it can augment the other classes by transferring the style of the images with diseased leaves onto the images with healthy leaves.
The cycle-generative adversarial network (CycleGAN) model \cite{cyclegan} has been proven to be effective for style transfer tasks \cite{liu2020unity}. Also, CycleGAN is a good fit for the augmentation task since it does not require paired inputs and works just as well with unpaired inputs, which is the case for the augmentation task.

In addition to learning the mapping from the source domain $X$ to the target domain $Y$, CycleGAN also learns the inverse mapping from the target domain to the source domain \cite{cyclegan}. In this sense, two generators and two discriminators are used, one of each type for each mapping: $G$ and $D_Y$ for the direct mapping $G : X \rightarrow Y$, as well as $F$ and $D_X$ for the inverse mapping $F : Y \rightarrow X$. Furthermore, CycleGAN introduces another objective, called cycle consistency loss, in addition to the classic GAN objectives \cite{goodfellow2014generative}. Intuitively, this novel objective uses the L1 norm to enforce the consistency of the inverse mapping, canceling the direct mapping, as shown in Eq. (\ref{eq:cycle_loss}).

\begin{multline}
    \mathcal{L}_{cyc}(G,F)=\mathbb{E}_{x \sim p_{data}(x)}[||F(G(x)) - x||_1] + \\ +\mathbb{E}_{y \sim p_{data}(y)}[||G(F(y)) - y||_1]
    \label{eq:cycle_loss}
\end{multline}

When passing an image from $X$ through generator $F$ or an image from $Y$ through generator $G$, the output is expected to be the same provided image since the input already comes from the output distribution. Therefore, a second regularization loss called the identity loss is used to constrain the model. This loss uses the same L1 norm as the cycle consistency loss and can be expressed through Eq. (\ref{eq:id_loss}).
\begin{multline}
    \mathcal{L}_{identity}(G,F)=\mathbb{E}_{x \sim p_{data}(x)}[||F(x) - x||_1] + \\ + \mathbb{E}_{y \sim p_{data}(y)}[||G(y) - y||_1] 
    \label{eq:id_loss}
\end{multline}

By combining the identity and cycle consistency losses with the standard GAN objectives, the final loss becomes as described by Eq. (\ref{eq:cyclegan_loss}).
\begin{multline}
    \mathcal{L}(G,F,D_X,D_Y)=\mathcal{L}_{GAN}(G, D_Y, X, Y) + \\ + \mathcal{L}_{GAN}(F,D_X,Y,X) + \lambda_1 \mathcal{L}_{cyc}(G,F) + \\ + \lambda_2 \mathcal{L}_{identity}(G,F)
    \label{eq:cyclegan_loss}
\end{multline}

Therefore, we train the CycleGAN model on our dataset of segmented images for each combination of healthy and diseased classes. This means that four CycleGAN models were trained in total. After training, the models were used on every segmented image of the healthy class to generate its diseased counterpart. The dataset was augmented using those generated diseased images by supplementing every diseased class such that the number of samples in each class is equal. For this, the required number of images was randomly picked from the generated images.

\subsection{Online Augmentations}

During training, online augmentation of the dataset was also tested.  The most basic augmentations that have been tested are horizontal and vertical flips, as well as random rotations. Therefore, an image is rotated with a random angle between 0 and 180 degrees, after which random horizontal and vertical flips are applied with a probability of 25\% each.

More advanced techniques include MixUp \cite{mixup}, CutMix \cite{cutmix}, Cutout \cite{cutout}, and FMix \cite{fmix}.
These augmentations are applied during the batching process so that the training dataset will differ with each epoch.
For the batched augmentations, there is a 50\% probability that the augmentation will be applied; otherwise, the batch is left unmodified. Furthermore, all batched augmentations use a random parameter $\lambda$, sampled from the beta distribution each time the augmentation is applied. For FMix, the beta distribution's $\alpha$ and $\beta$ parameters are set to 1, while for the other batched augmentations, both parameters are set to 0.8. With CutMix and Cutout, the square that is cut out is chosen to have the center at least one-quarter away from the edge of the image, as most images have the leaf centered in the middle of the image.

\subsection{Classification Models}

After segmentation and augmentation, the dataset is ready for training classification models. Thus, Transformer-based architectures (i.e., ViT \cite{vit} and CvT \cite{cvt}) were tested using different hyperparameters, sizes, and augmentation techniques. In order to compare these models to convolutional state-of-the-art models, ResNet was tested in some scenarios.

\section{Experimental Setup}
\label{sec:experiments}

\subsection{Performance Metrics}

For model evaluation, macro-averaging was used to combine accuracy, precision, recall, and F1-score binary classification metrics into multiclass metrics.

Apart from the initial ViT and ResNet tests, all tests also feature the top-k accuracy metric. Top-k accuracy means that when calculating the accuracy score, a prediction is considered accurate if any of the top-k outputs with the highest confidence is correct. For testing, the top-2 accuracy was used in addition to the classic accuracy metric.

\subsection{Hyperparameters}

\subsubsection{Pix2pix Setup}

The pix2pix model was trained using the Adam optimizer \cite{adam2015diederik} with a learning rate of $2*10^{-4}$ and the momentum parameters $\beta_1$ and $\beta_2$ set to 0.5 and 0.999, respectively. These parameters are the same as in the pix2pix paper \cite{pix2pix}. Other parameters taken from the paper are the batch size of 1, a value shown to work best for the pix2pix model, and the L1 loss weight, $\lambda$, set to 100. For the discriminator, we used the  PatchGAN \cite {li2016combining, pix2pix} of output size 30x30. The model was trained on the whole dataset for 25 epochs.

\subsubsection{CycleGAN Setup}

For training the CycleGAN models, the Adam optimizer was used along with the same learning rate, momentum parameters, and batch size as for the pix2pix model. The weights for the cycle consistency loss and identity loss were also taken from the CycleGAN paper \cite{cyclegan}, specifically from the ``Monet paintings to photos'' experiment, and were set to 10 and 5, respectively. From an architectural point of view, the generator with nine residual blocks from the paper was used, along with the 70x70 PatchGAN discriminator, also from the paper \cite{cyclegan}. Each model was trained on the whole segmented dataset for 100 epochs.

\subsubsection{Classification Model Setup}

All models were trained using a batch size of 32, with a few exceptions where batch sizes of 16 were used due to limited resources. The Adam optimizer was used, along with a learning rate scheduler that multiplies the learning rate by 0.25 every 15 epochs.

\section{Quantitative Results}

\subsection{Results for Offline Augmentations}

Initially, we tested a ViT-small model with a patch size of 16 and an input size of 224. The tests aim to compare different models and augmentation techniques, so the first tests involve finding the optimum hyperparameters at which to train the following models. Therefore, the model is first tested with and without dropout and learning rates of 0.001 and 0.0002 for 50 epochs to observe how hyperparameters affect the scores and the evolution of train and dev losses.

\begin{table*}[!t]
\center
\caption{ViT-small scores for different combinations of hyperparameters.}
\resizebox{0.6\textwidth}{!}{
\begin{tabular}{ l | c | c | c | c | c | c | c}
\hline
\multicolumn{2}{c|}{\textbf{Dropout}}    & \textbf{Augmented} & \textbf{Learning rate} & \textbf{Accuracy} & \textbf{Precision} & \textbf{Recall} & \textbf{F1}   \\\hline
\multicolumn{2}{l|}{No}         & Yes       & 0.001         & 64.1     & 54.4      & 56.9   & 55.0 \\
\multicolumn{2}{l|}{No}         & No        & 0.001         & 51.9     & 32.3      & 32     & 32.1 \\
\multicolumn{2}{l|}{No}         & Yes       & 0.0002        & 73.1     & 58.5      & 54.9   & 55.7 \\
\multicolumn{2}{l|}{No}         & No        & 0.0002        & \textbf{75.0}     & \textbf{61.6}      & \textbf{58.3}   & \textbf{59.4} \\
\multicolumn{2}{l|}{Yes}        & Yes       & 0.001         & 57.1     & 42.3      & 43.5   & 42.0 \\
\multicolumn{2}{l|}{Yes}        & No        & 0.001         & 59.6     & 36.2      & 36.6   & 36.2 \\
\multicolumn{2}{l|}{Yes}        & Yes       & 0.0002        & 73.7     & 59.1      & 56.5   & 57.2 \\
\multicolumn{2}{l|}{Yes}        & No        & 0.0002        & 64.7     & 47.3      & 47.1   & 46.3 \\ \hline
\end{tabular}}
\label{table:vit-initial_tests-01-02}
\end{table*}

\textbf{The augmentations improve performance in almost all cases.} \quad As can be observed from Table \ref{table:vit-initial_tests-01-02}, the scores in most cases are higher when the model is trained on the augmented dataset than the non-augmented dataset. In the case of the fifth and sixth rows, the model has higher accuracy when trained on the non-augmented dataset, but all the other scores are low. The explanation is that the model overfits the most frequent class because of the imbalanced nature of the non-augmented dataset. However, for the third and fourth rows, all the scores are higher for the non-augmented model than all the others. As the differences are not that big, this anomaly can be explained by the small dimension of the test set.

\textbf{The ViT-small model performs better when trained using a lower learning rate.} \quad In Table \ref{table:vit-initial_tests-01-02}, in the case of no dropout and a learning rate of 0.001, the model presents worse performance on both augmented and non-augmented datasets compared to the cases when a learning rate of 0.0002 is used. This means the lower learning rate is the correct choice, as the model presents better learning abilities regardless of the dataset. Another observation in our tests is that introducing dropout reduces overfitting mostly on the augmented dataset, with the losses on the non-augmented dataset remaining almost unchanged. Using a learning rate of 0.0002 and enabling dropout, the overfitting on the augmented dataset almost disappears, while it is greatly improved on the non-augmented dataset.

The ResNet-50V2 and ResNet-101V2 models were tested to compare how augmentation affects traditional convolutional neural networks. Like the ViT, these models are pre-trained on the ImageNet datasets \cite{imagenet}. A learning rate of 0.001 was used to test both augmented and non-augmented datasets, while all other settings were kept as before.

\begin{table*}[!htp]
\center
\caption{ResNet scores for different variants and augmentations.}
\resizebox{0.6\textwidth}{!}{
\begin{tabular}{ l | c | c | c | c | c | c}
\hline
\multicolumn{2}{c|}{\textbf{Model}}    & \textbf{Augmented} & \textbf{Accuracy} & \textbf{Precision} & \textbf{Recall} & \textbf{F1}   \\ \hline
\multicolumn{2}{l|}{ResNet-50V2}       & Yes            & \textbf{69.2}     & \textbf{48.7}      & \textbf{47.9}   & \textbf{47.7} \\
\multicolumn{2}{l|}{ResNet-50V2}       & No             & 62.8     & 43.8      & 44.5   & 43.9 \\
\multicolumn{2}{l|}{ResNet-101V2}      & Yes            & 61.5     & 43.6      & 42.3   & 42.6 \\
\multicolumn{2}{l|}{ResNet-101V2}      & No             & 48.7     & 31.9      & 31.3   & 31.5 \\ \hline
\end{tabular}}
\label{table:resnet-tests}
\end{table*}

\textbf{Bigger models encourage overfitting.} \quad In Table \ref{table:resnet-tests}, it can be noticed how ResNet performs worse in the case of the bigger model, as the higher number of parameters worsens overfitting. Furthermore, performance is better on the augmented dataset for both versions of ResNet, as the larger dataset discourages overfitting.

\textbf{ViT has better performance than ResNet.}  \quad When comparing the scores of the ViT-small with enabled dropout, augmentation, and a learning rate of 0.0002 to the ResNet-50V2 model with augmentation, it can be observed that the Transformer-based model performs better compared to the convolutional-based model.

By analyzing the loss evolution of the models, it could be noticed how the models started overfitting mostly around 20 to 30 epochs. For this reason, all tests described from now on will be done using 25 epochs. Regarding the other hyperparameters, enabled dropout, a learning rate of 0.0002, and the augmented dataset will be used. Until another model is tested, the ViT-small model will be used.

In order to further test how well the synthetically generated data captures the distribution of the real data, we compared the Train on Real - Test on Real (TRTR), Train on Real - Test on Synthetic (TRTS), Train on Synthetic - Test on Real (TSTR), and Train on Synthetic - Test on Synthetic (TSTS) scenarios \cite{trtr}. As the names suggest, these methods train and test the model on combinations of either the original dataset with real images or a dataset consisting of only synthetic images with no real images in the augmented classes. 
The synthetic data was thus extracted and split into train, dev, and test using the same ratios as before.

\begin{table*}[!htp]
\center
\caption{ViT-small scores for TRTR, TRTS, TSTR, and TSTS.}
\resizebox{0.6\textwidth}{!}{
\begin{tabular}{l | c | c | c | c | c | c}
\hline    
\multicolumn{2}{c|}{\textbf{Method}}     & \textbf{Accuracy}    & \textbf{Top-2 Accuracy} & \textbf{Precision} & \textbf{Recall} & \textbf{F1}   \\\hline
\multicolumn{2}{l|}{TRTR}   & 62.8     & 86.5           & 45.9      & 45.7   & 44.1 \\
\multicolumn{2}{l|}{TRTS}   & 62.3     & 84.0           & 60.1      & 55.7   & 55.4 \\
\multicolumn{2}{l|}{TSTR}   & 55.1     & 76.3           & 34.2      & 30.6   & 29.0 \\
\multicolumn{2}{l|}{TSTS}   & \textbf{96.9}     & \textbf{99.6}           & \textbf{97.1}      & \textbf{96.2}   & \textbf{96.6} \\\hline 
\end{tabular}}
\label{table:tsts}
\end{table*}

\textbf{Synthetic data poorly captures the distribution of the real data.} \quad Table \ref{table:tsts} shows how, to a certain extent, the synthetic data captures the distribution of the real data rather poorly. The discrepancy between TRTR and TSTS especially shows this. The model performs much better on synthetic data and learns by overfitting the specific properties of each generated class. As there is still one real class among the synthetic images, the healthy class, as well as all other synthetic classes being generated with different trained instances of CycleGAN, the model learns to distinguish between the classes by learning the different footprints left in the generated images.

\textbf{Models trained only on synthetic data generalize poorly to real data.} \quad The slight difference between TRTR and TRTS shows how a model trained on real data generalizes just as well to synthetic data, as the model learns the actual distribution without overfitting. The poor performance of TSTR compared to TSTS further emphasizes how the model overfits specific properties in the synthetic data and does not generalize well to real data.

\begin{table*}[!htp]
\center
\caption{ViT-small scores for different online augmentations.}
\resizebox{0.6\textwidth}{!}{
\begin{tabular}{l | c | c | c | c | c | c}
\hline    
\multicolumn{2}{c|}{\textbf{Augmentation}}     & \textbf{Accuracy}    & \textbf{Top-2 Accuracy} & \textbf{Precision} & \textbf{Recall} & \textbf{F1}   \\\hline
\multicolumn{2}{l|}{Rotation+Flips}        & 73.7     & 91.7           & 68.3      & 65.1   & 62.7 \\
\multicolumn{2}{l|}{MixUp}                 & 73.7     & 88.5           & 60.1      & 57.4   & 57.0 \\
\multicolumn{2}{l|}{CutMix}                & 76.3     & 88.5           & 62.9      & 59.9   & 59.7 \\
\multicolumn{2}{l|}{Cutout}                & 69.2     & 89.1           & 53.7      & 51.6   & 51.3 \\
\multicolumn{2}{l|}{FMix}                  & 74.4     & 89.7           & 64.2      & 57.8   & 59.1 \\
\multicolumn{2}{l|}{Rotation+Flips+MixUp}  & 69.9     & \textbf{93.6}           & 52.2      & 50.1   & 49.4 \\
\multicolumn{2}{l|}{Rotation+Flips+CutMix} & 73.7     & 88.5           & 66.4      & 57.5   & 59.3 \\
\multicolumn{2}{l|}{Rotation+Flips+Cutout} & 76.3     & 92.3           & 68.1      & 65.5   & 63.3 \\
\multicolumn{2}{l|}{Rotation+Flips+FMix}   & \textbf{77.6}     & 91.0           & \textbf{71.6}      & \textbf{67.1}   & \textbf{67.7} \\\hline
\end{tabular}}
\label{table:online-aug-tests}
\end{table*}

\subsection{Results for Online Augmentations}

\textbf{Online augmentations improve performance in almost all cases.} \quad It can be observed in Table \ref{table:online-aug-tests} how, with a few exceptions, online augmentations further improve the performance when compared to the equivalent configuration in Table \ref{table:vit-initial_tests-01-02}, specifically the second to last row. One of the exceptions is Cutout, which has no other augmentations.

\textbf{Cutout with no augmentations offers the worst performance.} \quad Because the segmented images have most pixels set to 0, cutting out additional parts of the image erases essential information. Adding other augmentations seems to counteract this, as the Cutout model with random rotations and flips is among the better-performing models.

\textbf{FMix features the best all-around performance.} \quad CutMix and FMix have similar performances, with FMix taking the lead, especially when adding rotations and flips. Rotations and flips seem to improve all metrics except accuracy, while batched augmentations have more of an effect on the standard accuracy metric.

Finally, we tested the CvT model \cite{cvt} on the augmented dataset, non-augmented dataset, and the various combinations of online augmentations also tested up until now. By analyzing the loss evolution, we observed that CvT is more prone to overfitting, being a bigger model than ViT. The augmented dataset helped with overfitting, as did the augmentations with random rotations and flips. The choice of batched augmentation did not impact overfitting much, but CutMix and FMix had slightly better performance in that aspect.

\begin{table*}[!htp]
\center
\caption{CvT scores for the offline augmented dataset (first row), the non-augmented dataset (second row), and the augmented dataset with combinations of online augmentations (from third to last row).}
\resizebox{0.6\textwidth}{!}{
\begin{tabular}{l | c | c | c | c | c | c}
\hline    
\multicolumn{2}{c}{\textbf{Augmentation}}     & \textbf{Accuracy}    & \textbf{Top-2 Accuracy} & \textbf{Precision} & \textbf{Recall} & \textbf{F1}   \\\hline
\multicolumn{2}{l|}{Augmented}             & 76.3     & 91.7           & 62.1      & 58.9   & 60.2 \\
\multicolumn{2}{l|}{Non-augmented}         & 59.0     & 80.8           & 43.4      & 44.2   & 43.4 \\
\multicolumn{2}{l|}{MixUp}                 & 75.0     & 89.7           & 55.6      & 53.6   & 54.1 \\
\multicolumn{2}{l|}{CutMix}                & 71.2     & 84.6           & 55.9      & 54.3   & 54.8 \\
\multicolumn{2}{l|}{Cutout}                & 75.0     & \textbf{92.9}           & 61.1      & 60.6   & 60.6 \\
\multicolumn{2}{l|}{FMix}                  & 75.0     & 85.9           & 60.5      & 60.8   & 59.8 \\
\multicolumn{2}{l|}{Rotation+Flips+MixUp}  & 76.3     & 89.7           & 63.2      & 62.6   & 62.8 \\
\multicolumn{2}{l|}{Rotation+Flips+CutMix} & 75.0     & 91.0           & 58.9      & 60.4   & 58.9 \\
\multicolumn{2}{l|}{Rotation+Flips+Cutout} & \textbf{78.2}     & \textbf{92.9}           & \textbf{66.1}      & \textbf{64.3}   & \textbf{63.4} \\
\multicolumn{2}{l|}{Rotation+Flips+FMix}   & 76.9     & 91.0           & 61.5      & 63.0   & 61.0 \\ \hline
\end{tabular}}
\label{table:cvt-tests}
\end{table*}

\textbf{CvT has similar performance to ViT.} \quad As seen in Table \ref{table:cvt-tests}, the performances of CvT are similar to ViT, with rotation and flip augmentations making the model carry out better. In the case of CvT, the versions where Cutout is used have similar performance and perform better than the other batched augmentations.
While ViT favored CutMix over MixUp, the performances are flipped with CvT, such that CutMix outperforms MixUp.
FMix seems to be the most robust, with its performance having smaller fluctuations than the rest of the batched augmentations, regardless of the model or the other online augmentations used. Using the non-augmented dataset has the same performance impact on CvT as it had on ViT.

\section{Qualitative Results}

\subsection{Pix2pix Examples}

\begin{figure}[!ht]
\centering
\includegraphics[width=0.7\columnwidth]{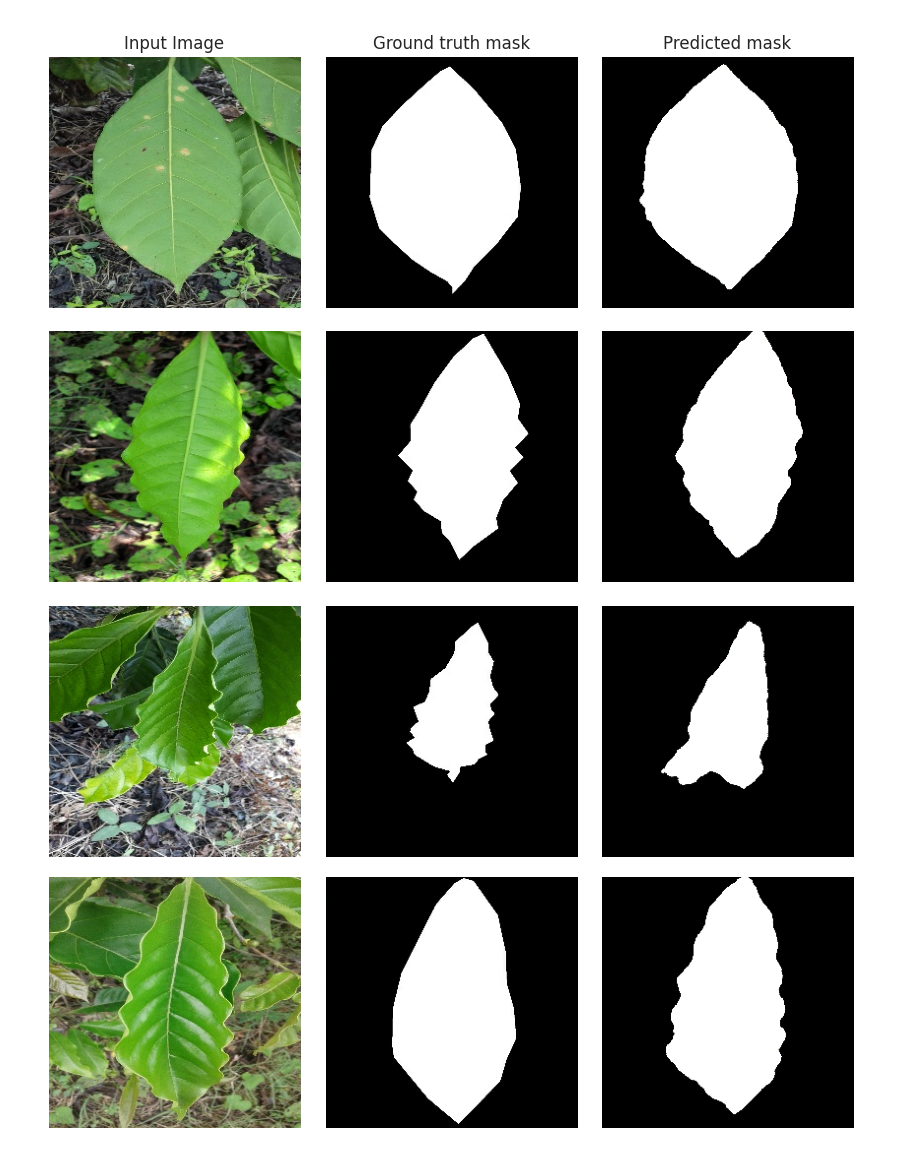}
\caption{Examples of pix2pix predicted masks compared to the ground truth masks.}
\label{fig:pix2pix_examples}
\end{figure}

\textbf{Predicted masks are smoother than hand-drawn masks.} \quad As can be seen in the first and second examples of Figure \ref{fig:pix2pix_examples}, the predicted mask is smoother around the edges compared to the ground truth mask, albeit less close to the outline of the leaf. The fourth example shows that the model could generate a mask that surpasses the ground truth in the correct conditions.

\textbf{Predicted masks sometimes include background leaves.} \quad The third image shows how certain background leaves hinder the model's ability to find an outline that follows the exact shape of the leaf, outputting a smoother segmentation instead. Therefore, in areas where the outline is less visible, the model also includes the background leaves when performing segmentation.

\textbf{The model has trouble segmenting leaves against specific backgrounds.} \quad Because the pix2pix model was trained on images that mostly contain backgrounds of other leaves combined with grass and dirt, it has trouble segmenting images of leaves taken against backgrounds such as asphalt and concrete. Surprisingly, the model is good at segmenting images of leaves taken against plain or solid color backgrounds.

\subsection{CycleGAN Examples}

\begin{figure}[!ht]
\centering
\includegraphics[width=1.0\columnwidth]{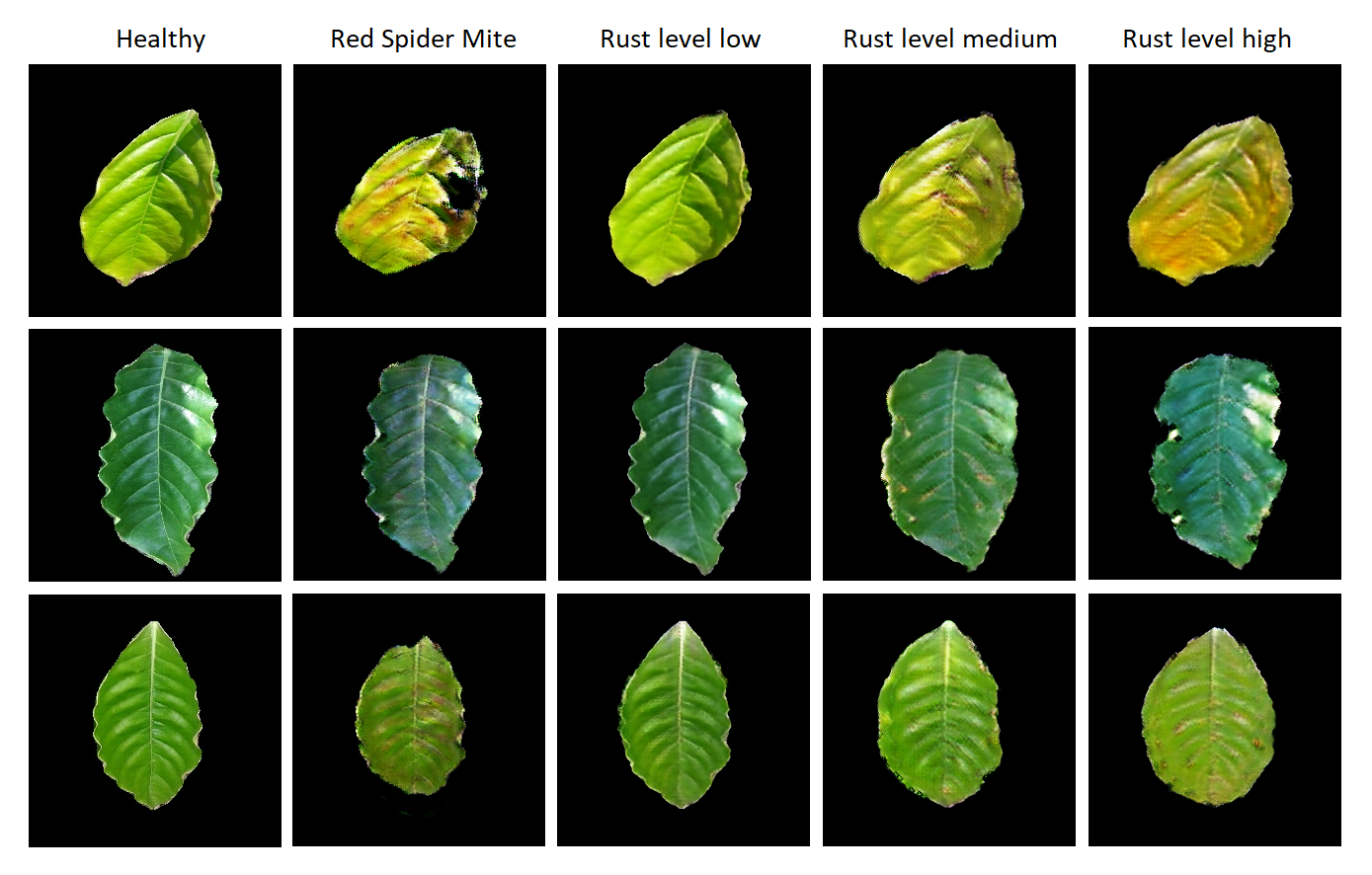}
\caption{Examples of diseased leaf images generated from healthy leaf images.}
\label{fig:cyclegan_examples}
\end{figure}

\textbf{Some synthetic images contain noisy holes.} \quad As shown in the first example of Figure \ref{fig:cyclegan_examples}, the generated red\_spider\_mite image features a big hole in the leaf. While the hole is an excellent example of leaf damage, there is also a lot of noise surrounding it, which decays the quality of the sample. Better-looking hole examples are found in the high rust-level image generated from the second healthy leaf, the synthetic image featuring small holes around the edges of the leaf.

\textbf{The low rust-level synthetic images feature almost no discernible modifications.} \quad For the low rust level, there is little discernible difference between the healthy and diseased images. This might be because this level of rust has a minimal impact on the appearance of the leaf, making it look almost identical to a healthy leaf in most cases. In comparison, the medium and high levels of rust have the most visible effects. As expected, the first and third examples show more severe rust stains for the high level than for the medium level, while the second example shows more holes. 

\subsection{T-SNE Visualizations}

In order to show how the generated diseased leaf images are close to the actual images in the input distribution space, latent features are first extracted from the images by passing them through a ViT. This vision Transformer is pre-trained on the ImageNet-21k and ImageNet-1k datasets \cite{imagenet}, so it should extract distinctive features out of the leaf images without prior knowledge about the input distribution. In order to visualize those latent features, the t-distributed stochastic neighbor embedding (t-SNE) dimensionality reduction technique \cite{hinton2002tsne} is used.

\begin{figure}[!ht]
\centering
\includegraphics[width=1.0\columnwidth]{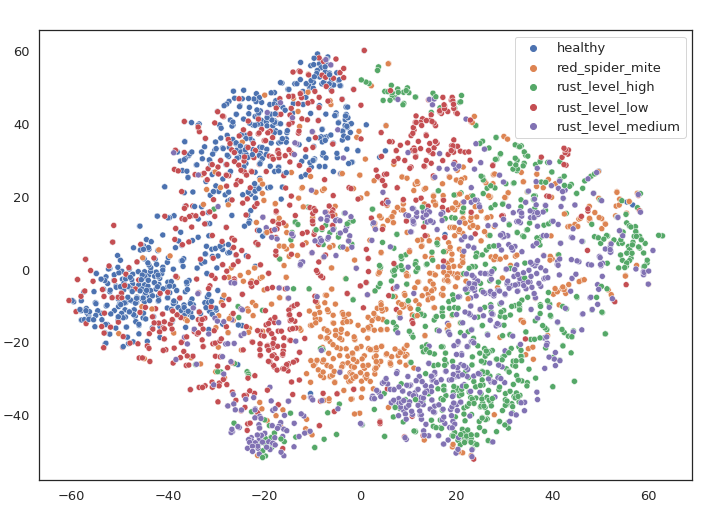}
\caption{2D t-SNE representations of each class's images, both synthetic and real.}
\label{fig:tsne_all}
\end{figure}

In Figure \ref{fig:tsne_all}, it is easy to notice how the different classes are each spread into their clusters, with the clusters also overlapping each other a bit. This is expected from a model pre-trained on a generic image dataset without prior knowledge of the plant leaf classification task. It can still be observed how there are no evident outliers for any class, with the synthetic images seemingly blending into the real images.

\begin{figure}[!ht]
\centering
\includegraphics[width=1.0\columnwidth]{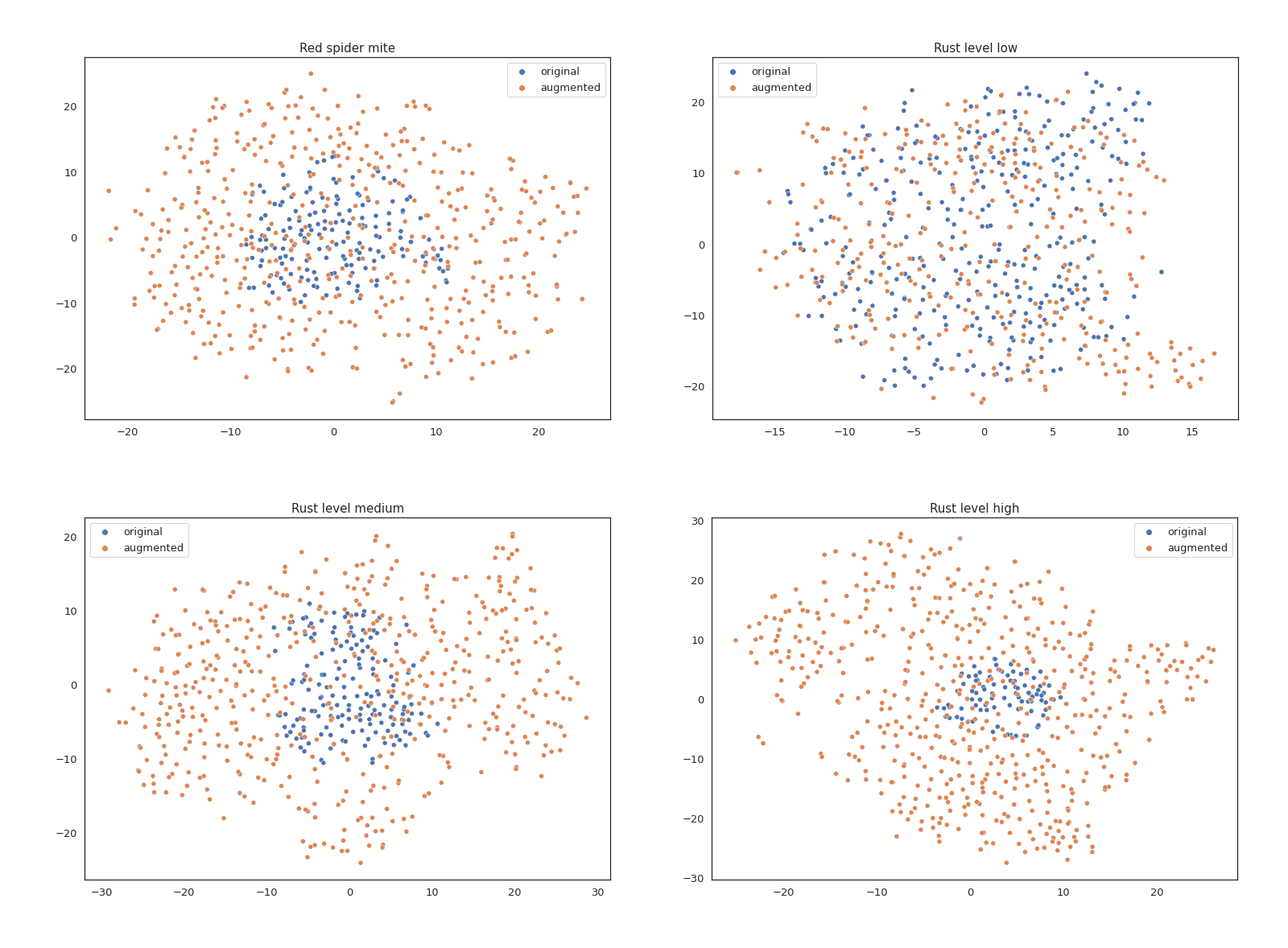}
\caption{2D t-SNE visualizations of each class, comparing real and synthetic images.}
\label{fig:tsne_each}
\end{figure}

In Figure \ref{fig:tsne_each}, for the most part, the synthetic image representations surround the real images representations, except for the red\_spider\_mite class, where the real and synthetic images overlap almost perfectly.

\subsection{CAM Visualizations}

\begin{figure}[!ht]
\centering
\includegraphics[width=1.0\columnwidth]{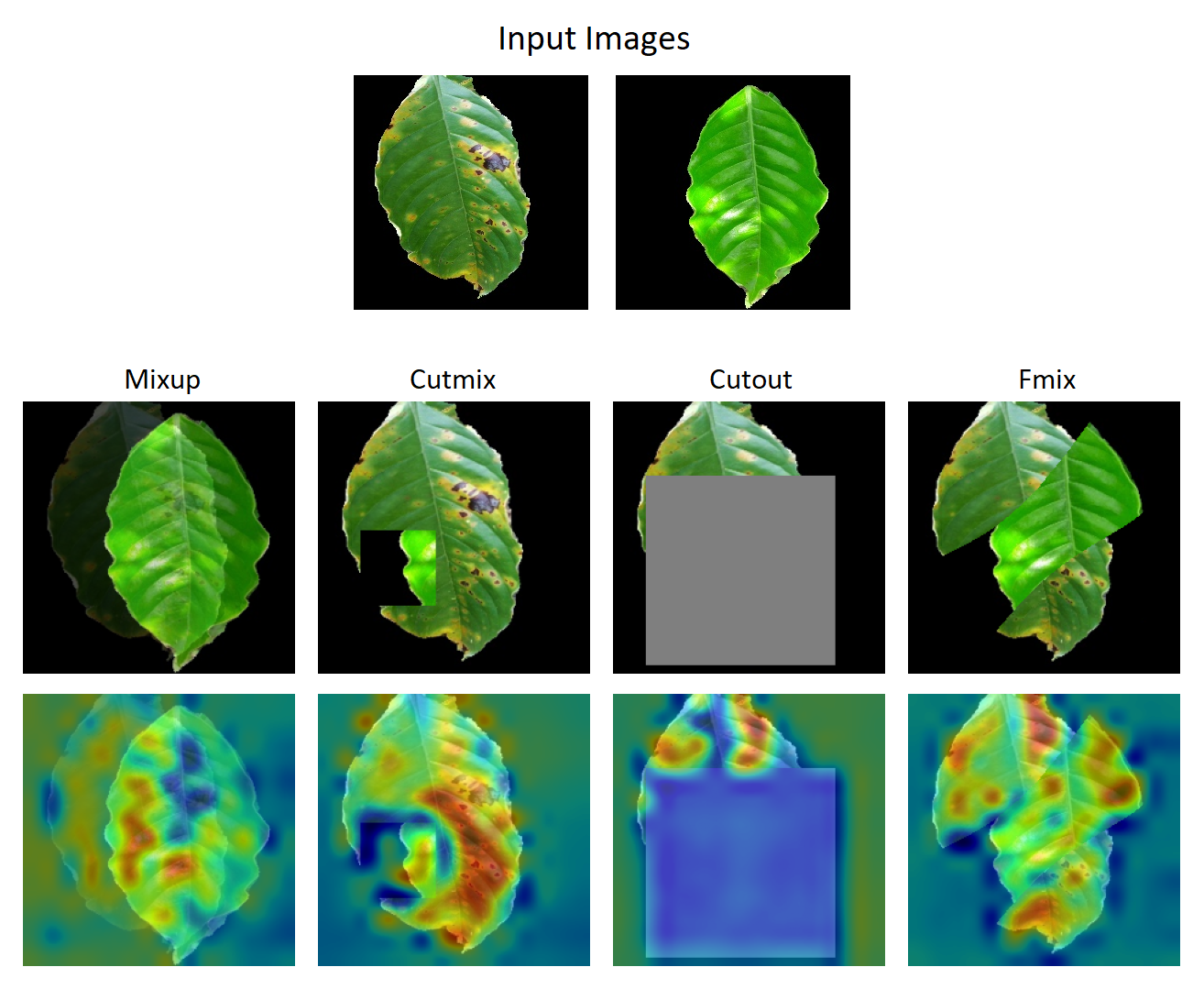}
\caption{CAM visualizations for batched augmentation techniques.}
\label{fig:cam-all_mixes}
\end{figure}

\textbf{Models trained with certain online augmentations focus better on the critical parts of leaves.} \quad  For the MixUp strategy, the model focuses more on the area where the two images overlap, and there are no rust spots (see Figure \ref{fig:cam-all_mixes}). Interestingly, the big rust region is mainly ignored, with the model focusing more on the small rust spots on the left side of the rusty leaf. For CutMix, the model learned to ignore the areas that encompass the transitions from one image to another while also focusing on the essential elements in both images. With Cutout, even when a big area is cut out in the image, the model learns to ignore it and focuses instead on the unobstructed regions. FMix CAM is similar to CutMix CAM, the model ignoring the transitions, especially in regions where one image transitions to the background of the other image.

\begin{figure}[!ht]
\centering
\includegraphics[width=1.0\columnwidth]{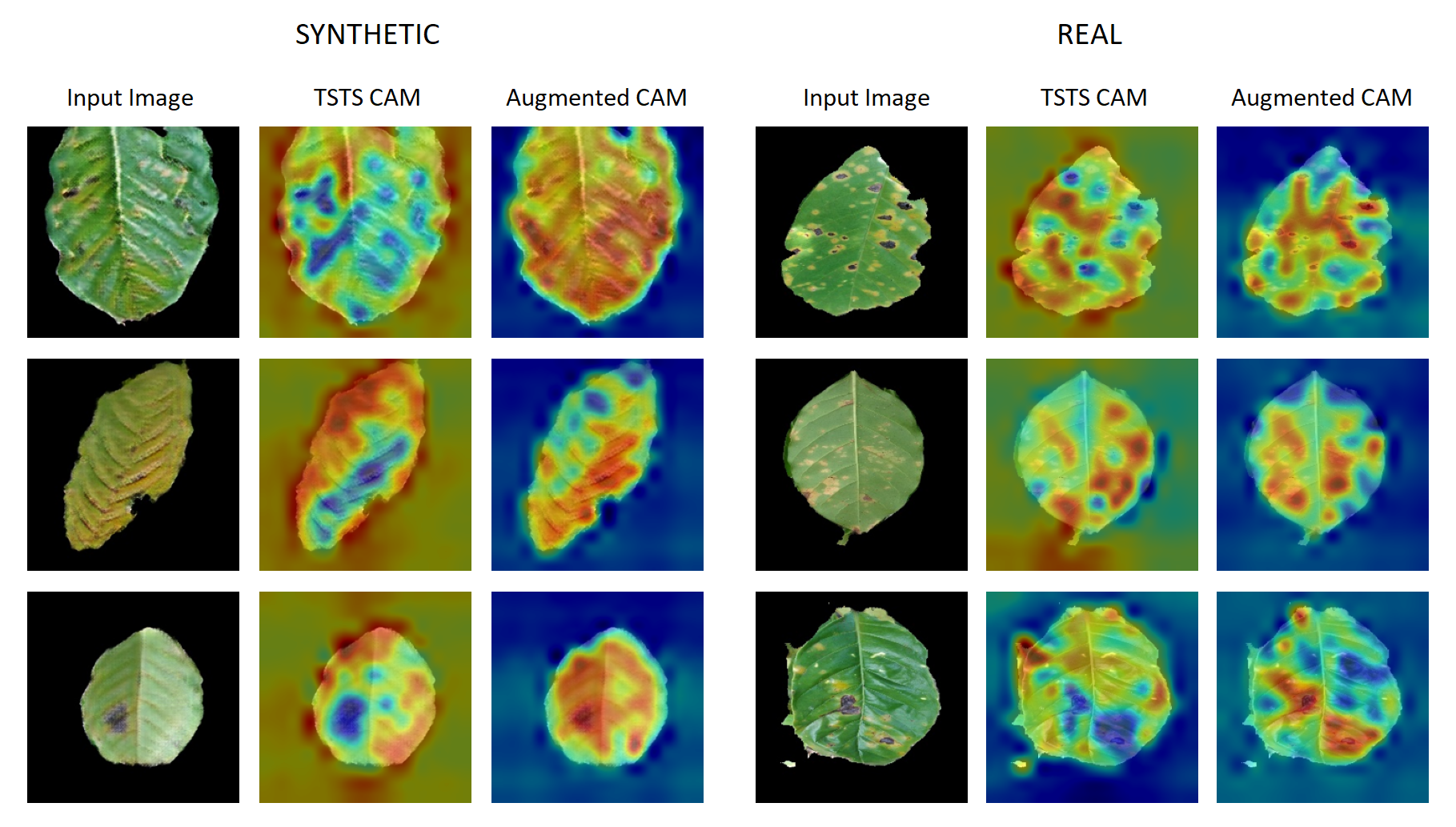}
\caption{CAM visualizations for synthetic (left) and real (right) for the TSTS model (TSTS CAM) and a model trained normally on the augmented dataset (Augmented CAM). All images are part of the rust\_level\_high class.}
\label{fig:cam-tsts_normal}
\end{figure}

\textbf{Models trained on synthetic data focus more on the background and edges of leaves.} \quad The CAMs in Figure \ref{fig:cam-tsts_normal} show how the model trained on the synthetic dataset focuses the most on the edges of the leaves, where more artifacts were left by the CycleGAN model when generating the images. Also, the TSTS model seems to focus surprisingly much on the background, which can be explained because the CycleGAN model leaves noise in the background of the generated image. The model trained on the augmented dataset focuses the most on areas where the leaf is affected by rust.

\textbf{Models trained using online augmentations focus on the image as a whole.} \quad By comparing the visualizations in Figure \ref{fig:cam-all_mixes} to the visualizations in Figure \ref{fig:cam-tsts_normal}, it can be observed that the models trained using online augmentations direct their attention more toward the background of the image compared to the model trained without online augmentations. This is because the models trained using online augmentations learn to focus on the image as a whole, not only on the center part.

\section{Conclusions and Future Work}
\label{sec:conclusion}

This paper presented a deep learning pipeline for the task of leaf disease classification on the RoCoLe dataset. Consequently, image leaf segmentation was performed using a pix2pix model, and CycleGAN was used to augment the diseased classes in the dataset. Also,  Transformer-based image classification models were trained on the augmented dataset using various online augmentations. 

This approach of using augmentations combined with Transformer models increased the classification performance compared to approaches using convolutional models or without augmentations. Even though the results of the TRTR, TRTS, TSTR, and TSTS scenarios showed that the synthetic data only vaguely captures the distribution of the real data, the increased performance of the models trained on the augmented dataset proved that the CycleGAN augmentations were helpful.

The presented approach can still be improved, with better alternatives being viable in most parts of the architecture. Therefore, the main improvement that can be implemented is the use of the StarGAN \cite{stargan} multi-domain image translation model for augmenting the RoCoLe dataset. This approach would benefit from needing only one model to be trained for augmenting all diseased classes. Other GAN variations can also be tested, such as the Wasserstein GAN \cite{wasserstein}, which has proven more stable than other GANs.

The leaf segmentation could also be improved by using semantic segmentation \cite{semantic_segmentation}. However, this would increase computational costs. A cheaper alternative would be to implement semantic segmentation using either the previously presented CAM method or the better GradCAM variant \cite{selvaraju2017gradcam}.

Finally, a Swin Transformer \cite{liu2021swin} can be used as a classifier instead of the ViT or CvT. Another alternative would be to use vision-language models \cite{coca}, which are the current state of the art in image classification. Models like the contrastive captioner \cite{coca} use encoder-decoder architectures to learn captions for images, so the encoder could be fine-tuned to classify images from the RoCoLe dataset. 

\section*{\uppercase{Acknowledgements}}
This work was supported by GNAC ARUT 2023.

\bibliographystyle{apalike}
{\small
\bibliography{example}}

\end{document}